\documentclass[10pt,twocolumn,twoside]{IEEEtran}

\usepackage{amsfonts}
\usepackage{amssymb}
\usepackage{amstext}
\usepackage{amsthm}
\usepackage{graphicx}
\usepackage{url}
\usepackage{epstopdf}
\usepackage{color}
\usepackage{xcolor}
\usepackage{bbm}
\usepackage[cmex10]{amsmath}
\usepackage{array}
\usepackage{caption}
\usepackage{lipsum,comment}
\usepackage{cite}
\usepackage{breqn}
\colorlet{mygreen}{green!60!gray}

\begin{document}
%
\title{Deep Learning for Resilient Adversarial Decision Fusion in Byzantine Networks}
%
%
%

\author{~Kassem Kallas,~\IEEEmembership{Senior Member,~IEEE}}
%

\maketitle

\begin{abstract}
This paper introduces a deep learning-based framework for resilient decision fusion in adversarial multi-sensor networks, providing a unified mathematical setup that encompasses diverse scenarios, including varying Byzantine node proportions, synchronized and unsynchronized attacks, unbalanced priors, adaptive strategies, and Markovian states. Unlike traditional methods, which depend on explicit parameter tuning and are limited by scenario-specific assumptions, the proposed approach employs a deep neural network trained on a globally constructed dataset to generalize across all cases without requiring adaptation. Extensive simulations validate the method's robustness, achieving superior accuracy, minimal error probability, and scalability compared to state-of-the-art techniques, while ensuring computational efficiency for real-time applications. This unified framework demonstrates the potential of deep learning to revolutionize decision fusion by addressing the challenges posed by Byzantine nodes in dynamic adversarial environments.
\end{abstract}


\IEEEpeerreviewmaketitle

\section{Introduction}
\label{sec:introduction}

Distributed decision fusion in adversarial environments has garnered significant attention, particularly in applications such as wireless sensor networks (WSNs) \cite{WSNDDByz, WSNanomalyDet}, cognitive radio networks (CRNs) \cite{SS2010, Rawat11, Zhang2013secure}, and multimedia forensics \cite{Bar13}. These systems, relying on data collected from multiple nodes, face threats from malicious nodes known as Byzantines~\cite{Lamp82}, which deliberately inject false data to compromise the system’s decision-making accuracy. Addressing Byzantine attacks is critical for improving robustness in collaborative and distributed systems, where reliable decision fusion is paramount \cite{Vemp13}.

In a distributed data fusion model, $n$ sensor nodes collect information about a system's state and communicate their observations (res. local decisions) to a central Fusion Center (FC). The FC aggregates the reports and makes a decision regarding the system’s state, represented as a binary vector $\mathbf{s} = \{s_1, s_2, \dots, s_m\}$ (See Figure~\ref{fig:setup}). However, with the presence of Byzantine nodes, traditional fusion rules, such as the Chair-Varshney rule \cite{OptFusion, varshney2012distributed}, fall short as they require assumptions about node reliability and attack characteristics that are often unknown in practical scenarios~\cite{MyBook}. Consequently, extensive research has been conducted to derive robust fusion strategies that can mitigate or resist Byzantine influence under diverse adversarial conditions \cite{Mar09, Rawat11, MyConsensus2016}.

Research into Byzantine-resilient decision fusion has evolved significantly, with studies exploring both classical and machine learning-based approaches. Initial efforts primarily focused on isolating Byzantines through probabilistic models and reputation-based mechanisms \cite{Vemp13, tolerant_scheme}. For instance, adaptive schemes like those in \cite{LearnByzantines} and isolation strategies in \cite{ConditionalFrequencyCheck} proposed frameworks for identifying and mitigating Byzantine nodes in CRNs (i.e. malicious secondary users), based on anomaly detection and collaborative learning. These approaches underscore the complexity of Byzantine behavior, highlighting the need for adaptive and context-sensitive fusion rules.

Recent research has also examined game-theoretic frameworks to address strategic interactions between Byzantine attackers and FCs. For example, \cite{MyTIFS} developed a zero-sum game model for optimum decision fusion rule derived using the Maximum A-Posteriori Probability (MAP), wherein the FC aims to maximize detection accuracy, while Byzantines minimize mutual information between their reports and the true state of the system. Building on this, \cite{MyInfoFusion2018} proposed a message-passing (MP) approach that reduced the computational complexity of MAP fusion in Markovian and i.i.d. environments. This work demonstrated that MP-based algorithms, leveraging Markovian dependencies, could perform near-optimally even with extensive observation windows, extending their applicability to real-time systems with complex state transitions.

Other studies have explored synchronization in adversarial attacks, where coordinated Byzantine actions increase the challenge of accurate fusion. For instance, \cite{MyMMEDIA2017} and \cite{MySynchronizedAttacksAPSIPA2018} analyzed scenarios where Byzantine nodes produce correlated reports, deceiving the FC by synchronizing their actions to closely mimic the statistics of legitimate nodes. Using factor graphs and MP, these studies achieved notable success in mitigating such attacks, yet the complexity of their solutions restricted scalability.

Machine learning techniques have recently been applied to enhance decision fusion's adaptability and robustness. For instance, \cite{ajay2022detection} explored neural networks for attacker detection in CRNs, while \cite{komar2016intelligent} used artificial neural networks optimized with immune plasma techniques for malicious behavior detection. Nevertheless, most of these machine learning approaches target detection rather than fusion, leaving room for improvement in computational efficiency and adaptability in fusion-based settings. In a related direction, \cite{behrens2023counter} discussed counter-adversarial resilience through ensemble-based fusion in networked systems, highlighting the growing need for fusion frameworks that are resilient yet scalable.

Despite these advances, optimal fusion strategies are often limited by computational demands, particularly when handling large networks, large observation window for the system, or highly dynamic attack strategies. To address this, \cite{MaxEntropyAttack} introduced maximum entropy attacks, demonstrating that high-entropy adversarial behavior could effectively degrade FC performance by increasing state uncertainty. Such findings emphasize on the importance of developing fusion methods that can handle high-entropy environments while maintaining computational tractability.

This paper makes several contributions to the problem of decision fusion in adversarial environments. First, it provides a unified mathematical approach that rigorously formulates the problem of decision fusion across diverse scenarios, including independent and identically distributed (i.i.d.) data, Markovian states, and both unsynchronized and synchronized Byzantine attacks, as well as under different assumptions about the behavior of Byzantine nodes. These include scenarios such as Unconstrained Maximum Entropy, Constrained Maximum Entropy, and Fixed Number of Byzantine nodes, which reflect a wide range of adversarial configurations. These formulations address the complexities inherent in adversarial decision fusion and establish a versatile foundation for robust decision-making across various attack strategies and network conditions.

In addition, the paper proposes a novel deep learning-based approach for fusion, where a neural network is trained to estimate the system states directly from the matrix of received reports. This eliminates the need for explicit assumptions about Byzantine node positions or behaviors, allowing the method to generalize across diverse scenarios. The effectiveness of this approach is validated through extensive experiments under varying adversarial conditions, including diverse attack strategies, Byzantine node proportions, adaptive behaviors, and unbalanced priors. The results consistently show that the proposed approach outperforms traditional methods in terms of accuracy, error probability, and robustness while maintaining computational efficiency for real-time applications. Collectively, these contributions provide a comprehensive and scalable solution for secure decision fusion in adversarial multi-sensor networks.

The rest of the paper is organized as follows. Section~\ref{sec.OptFus} presents a unified approach for adversarial decision fusion, deriving the mathematical formulations for various scenarios, including independent and identically distributed (i.i.d.) data, Markovian states, and assumptions about Byzantine behaviors such as Unconstrained Maximum Entropy, Constrained Maximum Entropy, and Fixed Number of Byzantine nodes. Section~\ref{sec.DL} introduces the proposed deep learning-based approach for decision fusion, detailing the neural network architecture, training methodology, and evaluation metrics. Section~\ref{sec.simul} provides an extensive experimental evaluation, demonstrating the robustness, scalability, and efficiency of the proposed method under diverse adversarial configurations. Section~\ref{sec.limitations} discusses the limitations of the proposed method. Finally, Section~\ref{sec.conc} concludes the paper with a summary of contributions, key findings, and potential directions for future research.

\section{Unified Derivation for Adversarial Decision Fusion Rules}
\label{sec.OptFus}
Consider a sensor network in which a Fusion Center (FC) collects reports from $n$ sensor nodes over an observation window of length $m$. The FC's objective is to determine the true state vector $\mathbf{s} = \{s_1, s_2, \ldots, s_m\}$, where each $s_i \in \{0, 1\}$ represents the system state at time $i$. Each sensor node $j$ provides a report $r_{i,j}$ for each state $s_i$, with nodes classified as either honest ($h_j = 1$) or Byzantine (malicious, $h_j = 0$). The FC aims to maximize the posterior probability $P(\mathbf{s} | \mathbf{R})$, where $\mathbf{R} = \{R_1, R_2, \ldots, R_m\}$ is the matrix of received reports, and $R_i = \{r_{i,1}, r_{i,2}, \ldots, r_{i,n}\}$ denotes the set of reports at time $i$. 

\begin{figure}[!t]
    \centering
    \includegraphics[width=\columnwidth]{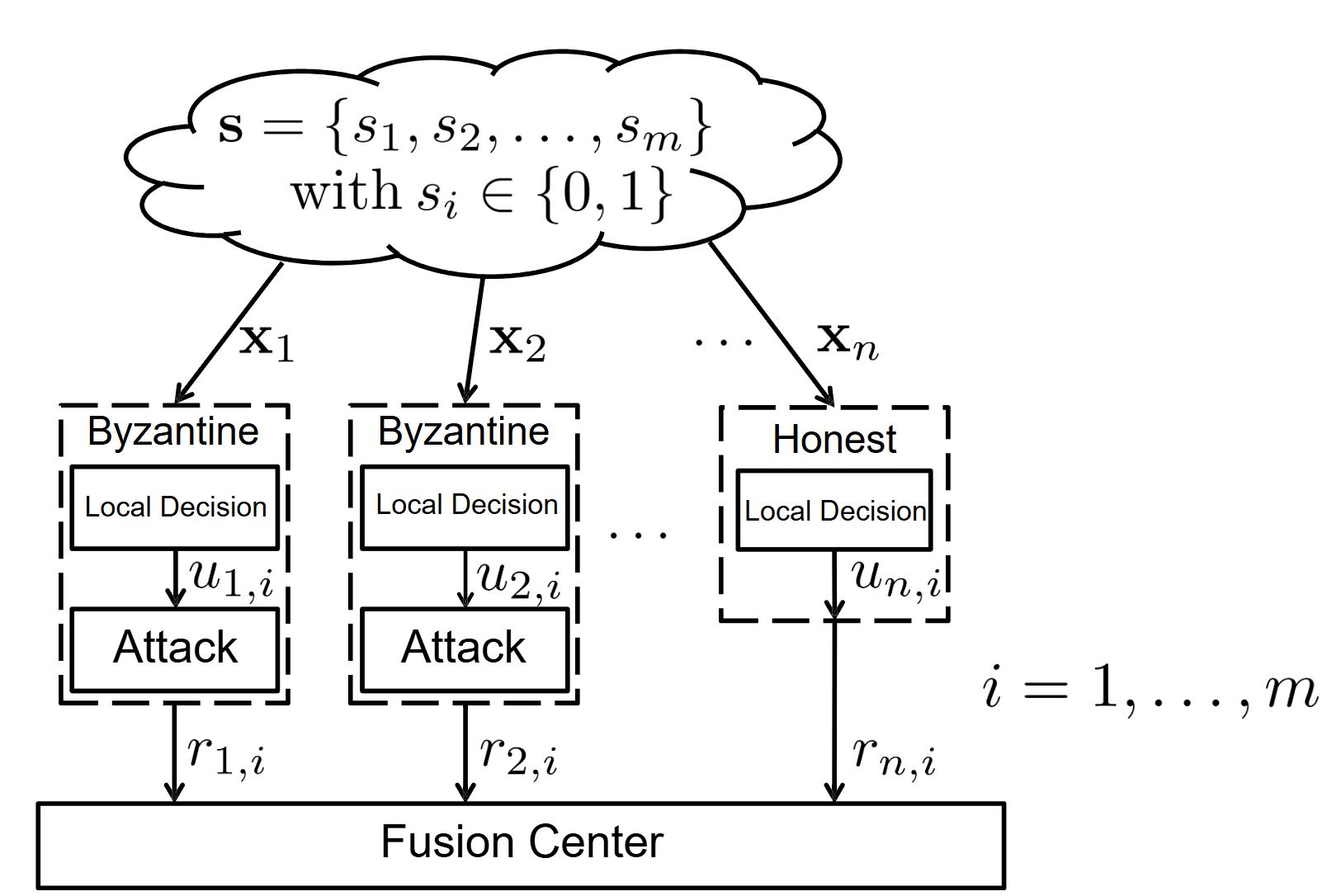} 
    \caption{Problem Setup.}
    \label{fig:setup}
\end{figure}

The problem setup, depicted in Figure~\ref{fig:setup}, shows the structure of this adversarial decision fusion scenario. Using Bayes' theorem, the posterior probability can be expressed as

\begin{equation}
P(\mathbf{s} | \mathbf{R}) = \frac{P(\mathbf{R} | \mathbf{s}) P(\mathbf{s})}{P(\mathbf{R})},
\end{equation}

where $P(\mathbf{R})$ is a normalizing constant independent of $\mathbf{s}$. Therefore, the FC maximizes the term $P(\mathbf{R} | \mathbf{s}) P(\mathbf{s})$.

For each state $s_i$, the likelihood of the reports $R_i$ given $s_i$ is given by:

\begin{equation}
P(R_i | s_i) = \prod_{j=1}^n P(r_{i,j} | s_i, h_j),
\end{equation}

where $P(r_{i,j} | s_i, h_j)$ represents the probability of receiving report $r_{i,j}$ from node $j$ given the state $s_i$ and the honesty of the node $h_j$. For honest nodes ($h_j = 1$), this probability can be expressed as:

\begin{equation}
P(r_{i,j} | s_i, h_j = 1) = (1 - \varepsilon) \delta(r_{i,j} - s_i) + \varepsilon (1 - \delta(r_{i,j} - s_i)),
\label{eq:honest_node_probability}
\end{equation}

where $\varepsilon$ is the error probability of an honest node, and $\delta$ is the Kronecker delta function. On the other hand, for Byzantine nodes ($h_j = 0$), the probability becomes:

\begin{equation}
P(r_{i,j} | s_i, h_j = 0) = (1 - P_{mal}) \delta(r_{i,j} - s_i) + P_{mal} (1 - \delta(r_{i,j} - s_i)),
\end{equation}

where $P_{mal}$ is the probability that a Byzantine node flips its report. By combining these probabilities, the overall likelihood is given by the following:

\begin{equation}
P(R_i | s_i) = \prod_{j=1}^n \left[ (1 - \alpha) P(r_{i,j} | s_i, h_j = 1) + \alpha P(r_{i,j} | s_i, h_j = 0) \right],
\end{equation}

where $\alpha$ denotes the fraction of Byzantine nodes in the network. The FC accounts for both honest and Byzantine nodes to maximize the likelihood of the true state.

Assuming equal prior probabilities for the states, i.e., $P(s_i = 0) = P(s_i = 1) = 0.5$, the likelihood term $P(R_i | s_i)$ can be expanded as:

\begin{equation}
\begin{aligned}
P(R_i | s_i) = \prod_{j=1}^n \Big[ & (1 - \alpha) \big( (1 - \varepsilon) \delta(r_{i,j} - s_i) \\
& + \varepsilon (1 - \delta(r_{i,j} - s_i)) \big) \\
& + \alpha \big( (1 - P_{mal}) \delta(r_{i,j} - s_i) \\
& + P_{mal} (1 - \delta(r_{i,j} - s_i)) \big) \Big].
\end{aligned}
\label{eq:likelihood}
\end{equation}

The FC then decides $s_i$ by maximizing $P(R_i | s_i)$.

To extend this framework to different cases, the i.i.d. assumption implies independence between observations. Depending on the subcase, the number of Byzantine nodes can vary. In the \textbf{Unconstrained Maximum Entropy Case}, the probability $\alpha$ is chosen to maximize the entropy of the node states, ensuring the most unpredictable distribution of Byzantine nodes. In the \textbf{Constrained Maximum Entropy Distributions}, several subcases are considered: \textbf{Maximum Entropy with Given $\alpha$}, where the expected number of Byzantine nodes is specified; \textbf{Maximum Entropy with $k<h$}, where the number of Byzantine nodes is constrained to be less than a variable $h$. Finally, in the \textbf{Fixed Number of Byzantines Case}, the number of Byzantine nodes is fixed at $k$, allowing for a controlled and consistent number of Byzantine nodes. For each subcase, the likelihood remains consistent, but the prior on $\alpha$ or the number of Byzantines varies~\cite{MyTIFS}.

In scenarios with unbalanced priors, we consider $P(s_i = 0) = P_0$ and $P(s_i = 1) = P_1$, where $P_0$ and $P_1$ are the prior probabilities of states 0 and 1, respectively. The posterior probability is given by:

\begin{equation}
P(s_i | R_i) = \frac{P(R_i | s_i) P(s_i)}{P(R_i)}.
\end{equation}

Here, the likelihood $P(R_i | s_i)$ retains the same expression as in the case of balanced priors in Equation~\ref{eq:likelihood}. However, in the presence of unbalanced priors, the posterior probability combines this likelihood with the prior probabilities $P(s_i) = P_0$ for $s_i = 0$ and $P(s_i) = P_1$ for $s_i = 1$. The Fusion Center (FC) then decides $s_i$ by maximizing $P(R_i | s_i) P(s_i)$, as elaborated in~\cite{MyMMEDIA2017}.

In the Markovian case, the state transitions follow a Markov process:

\begin{equation}
P(s_i | s_{i-1}) = \rho \delta(s_i - s_{i-1}) + (1 - \rho) (1 - \delta(s_i - s_{i-1})),
\end{equation}

where $\rho$ is the state transition probability. The posterior probability for the sequence of states is given by:

\begin{equation}
P(\mathbf{s} | \mathbf{R}) = P(s_1 | R_1) \prod_{i=2}^m P(s_i | s_{i-1}, R_i).
\end{equation}

By using Bayes' theorem iteratively, we obtain:

\begin{equation}
P(s_i | s_{i-1}, R_i) = \frac{P(R_i | s_i) P(s_i | s_{i-1})}{P(R_i)}.
\end{equation}

This approach is comprehensively detailed in \cite{MyInfoFusion2018}.

For Hidden Markovian models with synchronized attacks, Byzantine nodes generate a fake sequence $\hat{s}$ and synchronize their attacks:

\begin{equation}
P(R_i | s_i, \hat{s}_i) = \prod_{j=1}^n P(r_{i,j} | s_i, \hat{s}_i, h_j).
\end{equation}

For honest nodes ($h_j = 1$), the probability is described by Equation~\ref{eq:honest_node_probability}. For Byzantine nodes ($h_j = 0$), the probability is conditioned on the fake state $\hat{s}_i$ generated by the Byzantine nodes, rather than the true system state $s_i$. This behavior reflects the adversarial nature of Byzantine nodes, which craft their reports to align with $\hat{s}_i$ in an attempt to mislead the system. The corresponding probability is expressed as:

\begin{equation}
P(r_{i,j} | s_i, \hat{s}_i, h_j = 0) = (1 - \varepsilon) \delta(r_{i,j} - \hat{s}_i) + \varepsilon (1 - \delta(r_{i,j} - \hat{s}_i)).
\end{equation}

Although the left-hand side of the equation is conditioned on $s_i$, this reflects the Fusion Center's modeling of the overall scenario, where $s_i$ is assumed to be the true state. The right-hand side correctly represents the Byzantine nodes' adversarial strategy, which operates independently of $s_i$ and relies on $\hat{s}_i$ to generate misleading reports.

The posterior probability is then given by:

\begin{equation}
P(\mathbf{s} | \mathbf{R}, \hat{s}) = P(s_1 | R_1, \hat{s}_1) \prod_{i=2}^m P(s_i | s_{i-1}, R_i, \hat{s}_i).
\end{equation}

This scenario, including the adversarial strategies and their impact on the Fusion Center's decision-making, is thoroughly examined in \cite{MySynchronizedAttacksAPSIPA2018}.

\section{Deep Learning Approach}
\label{sec.DL}
In this section, we propose a deep learning approach to enhance decision fusion in the presence of Byzantine attacks. The idea is to design a neural network that takes the matrix of states $\mathbf{R}$ as input and outputs an estimated system state vector.

The proposed neural network is designed to take the matrix $\mathbf{R}$ of size $m \times n$ as input, where $m$ is the length of the observation window and $n$ is the number of sensor nodes. The network architecture, which will be specified in the experimental section, processes this input to produce the estimated state vector $\hat{\mathbf{s}}$, which corresponds to the system's state at each time step.

The network is trained using supervised learning, where the true state vector $\mathbf{s}$ serves as the ground truth. The loss function used is the mean squared error (MSE), defined as:
\begin{equation}
L(\mathbf{s}, \hat{\mathbf{s}}) = \frac{1}{m} \sum_{i=1}^m \left( s_i - \hat{s}_i \right)^2,
\end{equation}
where $\hat{s}_i$ is the estimated state at time step $i$.

To assess the performance of the proposed deep learning approach, we employed multiple evaluation metrics: mean squared error loss, error probability ($P_e$), bit error rate (BER), and accuracy. The error probability $P_e$ quantifies the likelihood that the estimated system state vector $\hat{\mathbf{s}}$ differs from the true state vector $\mathbf{s}$ and is defined as:
\begin{equation} \label{eq:Pe_theoretical}
P_e = P(\mathbf{s} \neq \hat{\mathbf{s}}),
\end{equation}
where $\mathbf{s}$ is the true system state vector and $\hat{\mathbf{s}}$ is the estimated vector. Empirically, $P_e$ is calculated as:
\begin{equation} \label{eq:Pe_empirical}
P_e = \frac{\text{Number of misclassified samples}}{\text{Total number of samples}}.
\end{equation}

The bit error rate (BER) measures the average error at the bit level within the system state vector and is defined as:
\begin{equation} \label{eq:BER_theoretical}
\text{BER} = \frac{1}{m} \sum_{i=1}^{m} P(s_i \neq \hat{s}_i),
\end{equation}
where $s_i$ and $\hat{s}_i$ are the $i$-th bits of $\mathbf{s}$ and $\hat{\mathbf{s}}$, respectively, and $m$ is the sequence length.

Lastly, the accuracy of the system is expressed as:
\begin{equation} \label{eq:accuracy}
\text{Accuracy} = 1 - P_e.
\end{equation}

\section{Simulation results and discussion}
\label{sec.simul}
\subsection{Simulation Setup}
We performed extensive simulations to assess the robustness of our proposed decision fusion model against Byzantine attacks across various configurations and adversarial scenarios. The simulations were conducted using controlled datasets, a custom DNN model architecture, and carefully chosen experiment parameters to provide a comprehensive evaluation of the proposed method.

\subsubsection{Dataset Generation and Experiment Overview}
The dataset generation process and experiments were designed to simulate challenging conditions while providing a comprehensive evaluation of the model's robustness. Each network comprised $n \in \{10, 20, \dots, 100\}$ nodes, incremented by 10, operating in either honest or Byzantine modes. The sequence length (observation window), $m$, varied across $\{5, 10, \dots, 50\}$ with a step size of 5, capturing diverse temporal dependencies. Samples per class were tested over $[50, 100, 250, 500, 1000, 2000]$ to evaluate the model's behavior under varying data volumes. A class is defined as a specific configuration with fixed parameters, such as the proportion of Byzantine nodes $\alpha$, the sequence length $m$, and the probability of adversarial actions $P_{mal}$. Using these parameters, multiple reports in $\mathbf{R}$ were generated, each corresponding to different system state vectors $\mathbf{s}$. By varying the number of samples per class, we analyzed the impact of data volume on the model's training efficiency and robustness.

The dataset was fully generated at the beginning of each of the experiments. Before training, it was split randomly entry-wise into $80\%$ for training and $20\%$ for testing, ensuring that the test set consisted of unseen data, independent of the training process.

Data generation incorporated both i.i.d and Markovian assumptions. i.i.d data was created by flipping bits with a fixed probability across time steps, while Markovian data relied on a two-state Markov chain with transition probabilities $\rho \in \{0.05, 0.15, \dots, 1.0\}$, incremented by $0.1$, to model time-correlated system states in $\mathbf{s}$ and, consequently, the generated reports in $\mathbf{R}$. To modulate adversarial influence, the proportion of Byzantine nodes $\alpha$ was gradually increased from $0\%$ to $100\%$. Adversarial actions, represented as the flipping of observations, were parameterized by $P_{mal}$, which ranged from moderate ($P_{mal} = 0.1$) to extreme ($P_{mal} = 1.0$), simulating attacks of increasing severity. Both synchronized and unsynchronized attack patterns were considered, differentiating between coordinated and independent adversarial behaviors. Throughout all configurations, the error probability of an honest node $\varepsilon$ was fixed to $0.1$, ensuring consistency in the data corruption process.

Additional variations included balanced and unbalanced state priors, where $P_0$ was set to 0.5 in balanced configurations and adjusted otherwise. Fixed and variable numbers of Byzantine nodes, as detailed in Section~\ref{sec.OptFus}, were tested to explore the model's robustness under static and dynamic adversarial configurations. Maximum entropy constraints were applied, ensuring $k < h$ for variable node distributions, further enhancing the evaluation framework.

\subsubsection{DNN Model Architecture}
The DNN used in our experiments incorporated seven fully connected layers, with the input layer matching the flattened size, $n \times m$. The first hidden layer consisted of 2048 neurons, and subsequent layers halved the number of neurons at each step, following the progression: 2048, 1024, 512, 256, 128, and 64 neurons. Each layer utilized ReLU activations for non-linearity, and batch normalization was applied after each layer to enhance stability during training. The output layer consisted of $m$ neurons and employed a sigmoid activation function, which outputs a probability for each time step, ensuring the network produces an estimated system state vector $\hat{\mathbf{s}}$. The model was trained using the mean squared error loss function and optimized with the Adam optimizer with default parameters, running for $150$ epochs with a batch size of $512$ per case. The datasets were constructed with a precision of \texttt{float32} to ensure numerical consistency. All simulations were conducted on a \textbf{Dell Precision 5820 Tower} computer with an \textbf{Intel(R) Xeon(R) W-2265 CPU @ 3.50GHz}, \textbf{32GB RAM}, and \textbf{NVIDIA GeForce RTX 3090} GPU with \textbf{24GB of dedicated memory}.

\subsection{Comprehensive Evaluation of Model Performance and Robustness}

\subsubsection{Performance Comparison with State-of-the-Art Methods}
To assess the performance of our proposed deep learning (DL) approach, we compared it with several state-of-the-art methods, including majority voting~\cite{OptFusion}, hard isolation of Byzantine nodes (\textbf{HardIS})~\cite{Rawat11}, soft isolation of Byzantine nodes (\textbf{SoftIS})~\cite{MyCDC}, and the optimum maximum a-posteriori probability rule (\textbf{OPT})~\cite{MyTIFS}. Notably, the results reported for \textbf{SoftIS} and \textbf{OPT} were computed at the Nash equilibrium of a game-theoretic framework between the attacker and defender. In contrast, our DL method does not rely on a predefined game-theoretic setup but is trained on a comprehensive dataset encompassing a variety of scenarios. This test dataset consists of samples generated independently from the training dataset and is never seen during training, ensuring unbiased evaluation.

\begin{table*}[htbp]
\centering
\caption{Error probability for various fusion schemes under the following settings: $m = 4$, $n = 20$, $\varepsilon = 0.1$}
\label{tab:soa-m4}
\begin{tabular}{|c|c|c|c|c|c|c|}
\hline
\textbf{} & \textbf{Maj}~\cite{OptFusion} & \textbf{HardIS}~\cite{Rawat11} & \textbf{SoftIS}~\cite{MyCDC} & \textbf{OPT}~\cite{MyTIFS} & \textbf{DL} & \textbf{DL-BER} \\
\hline
i.i.d, $\alpha=0.3$ & 0.073 & 0.048 & 0.041 & 0.035 & 0.0 & 0.0 \\
\hline
i.i.d, $\alpha=0.4$ & 0.239 & 0.211 & 0.201 & 0.192 & 0.0 & 0.0 \\
\hline
i.i.d, $\alpha=0.45$ & 0.362 & 0.344 & 0.338 & 0.331 & 0.0 & 0.0  \\
\hline
fixed, $k=6$ & 0.017 & 0.002 & 6.2e-4 & 3.8e-4 & 0.0 &  0.0\\
\hline
fixed, $k=8$ & 0.125 & 0.044 & 0.016 & 0.004 & 0.0 & 0.0 \\
\hline
fixed, $k=9$ & 0.279 & 0.186 & 0.125 & 0.055 & 0.0 & 0.0 \\
\hline
Max Entropy, $k< n/2$ & 0.154 & 0.086 & 0.052 & 0.021 & 0.003 & 0.001 \\
\hline
Max Entropy, $k< n/3$ & 0.0041 & 5e-4 & 2.15e-4 & 1.9e-4 & 0.001 & 0.0003  \\
\hline
\end{tabular}
\end{table*}

\begin{table*}[htbp]
\centering
\caption{Error probability for various fusion schemes under the following settings: $m = 10$, $n = 20$, $\varepsilon = 0.1$}
\label{tab:soa-m10}
\begin{tabular}{|c|c|c|c|c|c|c|}
\hline
\textbf{} & \textbf{Maj}~\cite{OptFusion} & \textbf{HardIS}~\cite{Rawat11} & \textbf{SoftIS}~\cite{MyCDC} & \textbf{OPT}~\cite{MyTIFS} & \textbf{DL} & \textbf{DL-BER} \\
\hline
i.i.d, $\alpha=0.3$ & 0.073 & 0.0364 & 0.0346 & 0.033 & 0.0 & 0.0 \\
\hline
i.i.d, $\alpha=0.4$ & 0.239 & 0.193 & 0.19 & 0.187 & 0.0 & 0.0 \\
\hline
i.i.d, $\alpha=0.45$ & 0.363 & 0.334 & 0.333 & 0.331 & 0.0 & 0.0 \\
\hline
fixed, $k=6$ & 0.016 & 1.53e-4 & 1.41e-4 & 1.22e-4 & 0.0 & 0.0 \\
\hline
fixed, $k=8$ & 0.126 & 0.0028 & 9.68e-4 & 4.13e-4 & 0.0 & 0.0 \\
\hline
fixed, $k=9$ & 0.279 & 0.0703 & 0.0372 & 1.58e-3 & 0.0 & 0.0 \\
\hline
Max Entropy, $k< n/2$ & 0.154 & 0.027 & 0.0141 & 6.8e-4 & 0.002 & 0.0003  \\
\hline
Max Entropy, $k< n/3$ & 0.0039 & 9.8e-5 & 7.4e-5 & 5e-5 & 0.0055 & 0.0015 \\
\hline
\end{tabular}
\end{table*}

The experiments were conducted under the following settings: $n=20$, $m \in \{4, 10\}$, $\varepsilon=0.1$, $P_0=0.5$, $P_{mal}=1.0$, and $\alpha \in \{0.0, \dots, 1.0\}$ with a step of $0.05$. For each $\alpha$ value, a dataset containing $200$ samples was generated. Tables~\ref{tab:soa-m4} and~\ref{tab:soa-m10} present the results for $m=4$ and $m=10$, respectively, highlighting the superior performance of our DL method. For instance, with $m=4$, the DL method achieves $P_e = 0.0$ across i.i.d and fixed Byzantine scenarios, and significant gains are observed even in maximum entropy settings with $k<n/2$. Similar observations hold for $m=10$, as shown in Table~\ref{tab:soa-m10}.

The superior performance of our method stems from the capability of deep learning to learn complex patterns and correlations in the data without explicit parameter tuning. Unlike the MAP rule, which requires precise knowledge of system parameters, the DL model generalizes effectively across diverse scenarios and leverages higher-order dependencies from the report matrix $\mathbf{R}$, achieving consistently high accuracy across varying conditions.

\subsubsection{Evaluation of i.i.d and Markovian Data}

\begin{figure}[!t]
    \centering
    \includegraphics[width=\columnwidth]{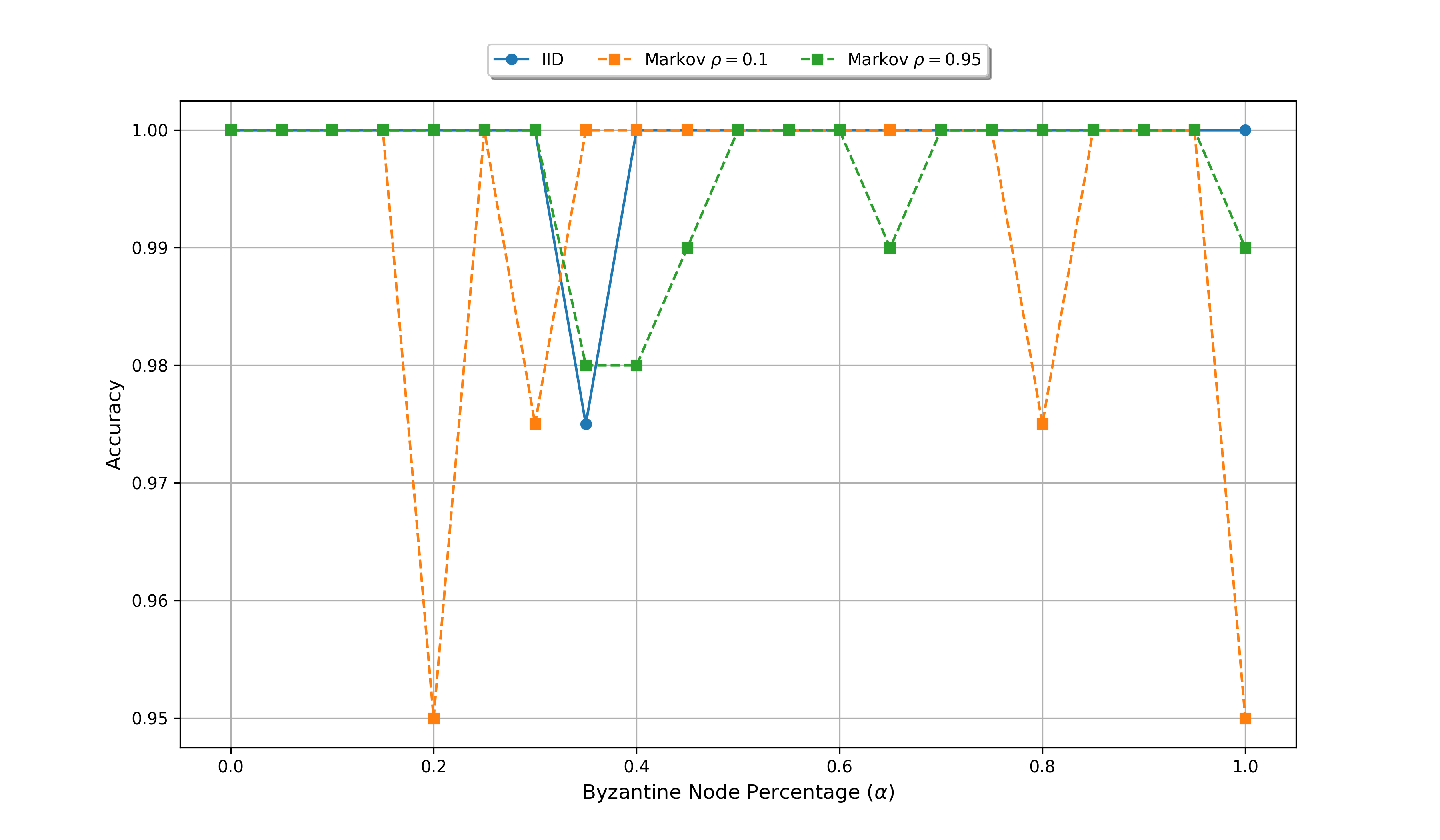}
    \caption{Model accuracy vs. $\alpha$ for i.i.d and Markovian data.}
    \label{fig:acc_vs_alpha}
\end{figure}

Figure~\ref{fig:acc_vs_alpha} illustrates the performance of our model under i.i.d and Markovian data for varying Byzantine percentages $\alpha \in \{0.0, \dots, 1.0\}$ with a step of $0.05$. The settings used were $n=20$, $m=4$, $\varepsilon=0.1$, $P_{mal}=1.0$, $P_0=0.5$ for the i.i.d case, and transition probabilities $\rho \in \{0.1, 0.95\}$ for the Markovian case. For each scenario, $200$ samples were generated for every $\alpha$ value. The results demonstrate that the DL model is highly robust across i.i.d and Markovian settings, achieving a minimum accuracy of $0.95$ ($P_e = 0.05$) as in the scenario of $\rho=0.1$ and $\alpha=0.2$ or $1.0$. Remarkably, in scenarios with $\alpha > 0.5$, where traditional fusion rules fail, the DL model adapts by learning to flip the global decision dominated by Byzantines, proofing its flexibility and resilience.

\subsubsection{Performance under Different Adversarial Scenarios}
\begin{figure}[!t]
    \centering
    \includegraphics[width=\columnwidth]{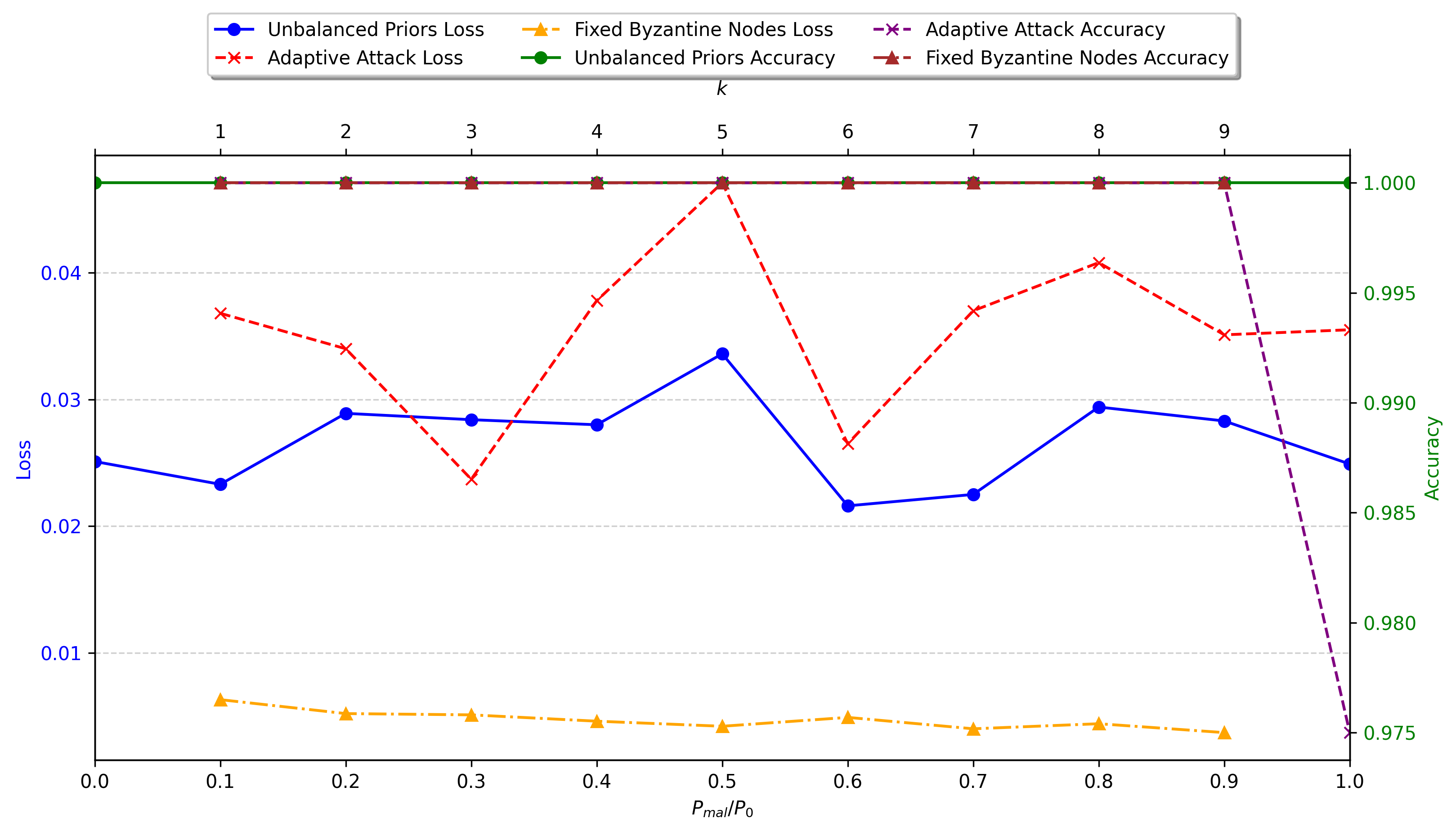}
    \caption{Performance comparison for Unbalanced Priors, Adaptive Strategies, and Fixed Byzantine Nodes.}
    \label{fig:performance_three_cases}
\end{figure}

In Figure~\ref{fig:performance_three_cases}, we evaluated the model’s performance across three distinct adversarial scenarios: 
\begin{itemize}
    \item The i.i.d case with unbalanced priors for varying $P_0$.
    \item The case of fixed Byzantine nodes with different $k$.
    \item The adaptive Byzantine case under various $P_{mal}$.
\end{itemize}
The experimental settings included $n=20$, $m=4$, $\varepsilon=0.1$, and $200$ samples per class. For the unbalanced priors scenario, $\alpha$ was fixed at $0.4$ and $P_{mal}$ at $0.1$, with the same $P_{mal}$ applied in the fixed Byzantine node scenario. For the adaptive Byzantine case, $P_0$ was fixed at $0.5$.

Across all scenarios, the plots reveal that the deep learning (DL) models consistently achieved an accuracy close to or equal to $1$ (or equivalently, $P_e$ close to or equal to $0$). The maximum observed loss for the mean square error objective function was below $0.05$. The worst performance was recorded in the adaptive Byzantine strategy with $P_{mal} = 1.0$, where the model achieved an accuracy of $0.975$. These results underline the robustness of the proposed DL method even under challenging adversarial settings.

\subsubsection{Global Dataset: Comprehensive Adversarial Settings}
To evaluate the robustness of our model against a wide range of adversarial settings, we constructed a global dataset that aggregates samples from various configurations, including varying Byzantine node percentages $\alpha \in \{0.0, \dots, 1.0\}$, synchronized and unsynchronized i.i.d data, unbalanced priors, adaptive and fixed Byzantine attack strategies with $k \in \{1, \dots, n\}$, maximum entropy with $k<h$ with $h \in \{n/4, n/3, n/2\}$, and Markovian data with different transition probabilities $\rho \in \{0.0, \dots, 1.0\}$. The dataset includes $200$ samples per configuration, enabling a unified evaluation framework. This approach eliminates the need to identify specific adversarial environments beforehand, making it practical for real-world applications.

Using the global dataset, we conducted experiments under the following settings: $n = 20$, $m = 4$, $\varepsilon = 0.1$, and $P_0 = 0.5$. The model demonstrated remarkable performance, achieving a minimal loss of $0.0001$, an accuracy of $0.9997$, an error probability of $0.0003$, and a bit error rate (BER) of $0.0002$. These results highlight the model's robustness across diverse adversarial scenarios, demonstrating that a single model trained simultaneously on all cases can perform effectively. This approach eliminates the need to train separate models for each adversarial scenario and removes the requirement of prior knowledge about the specific adversarial environment in which the model operates—an often infeasible or complex task.

\subsubsection{Impact of Varying Observation Window and Network Size}

\begin{figure}[!t]
    \centering
    \includegraphics[width=\columnwidth]{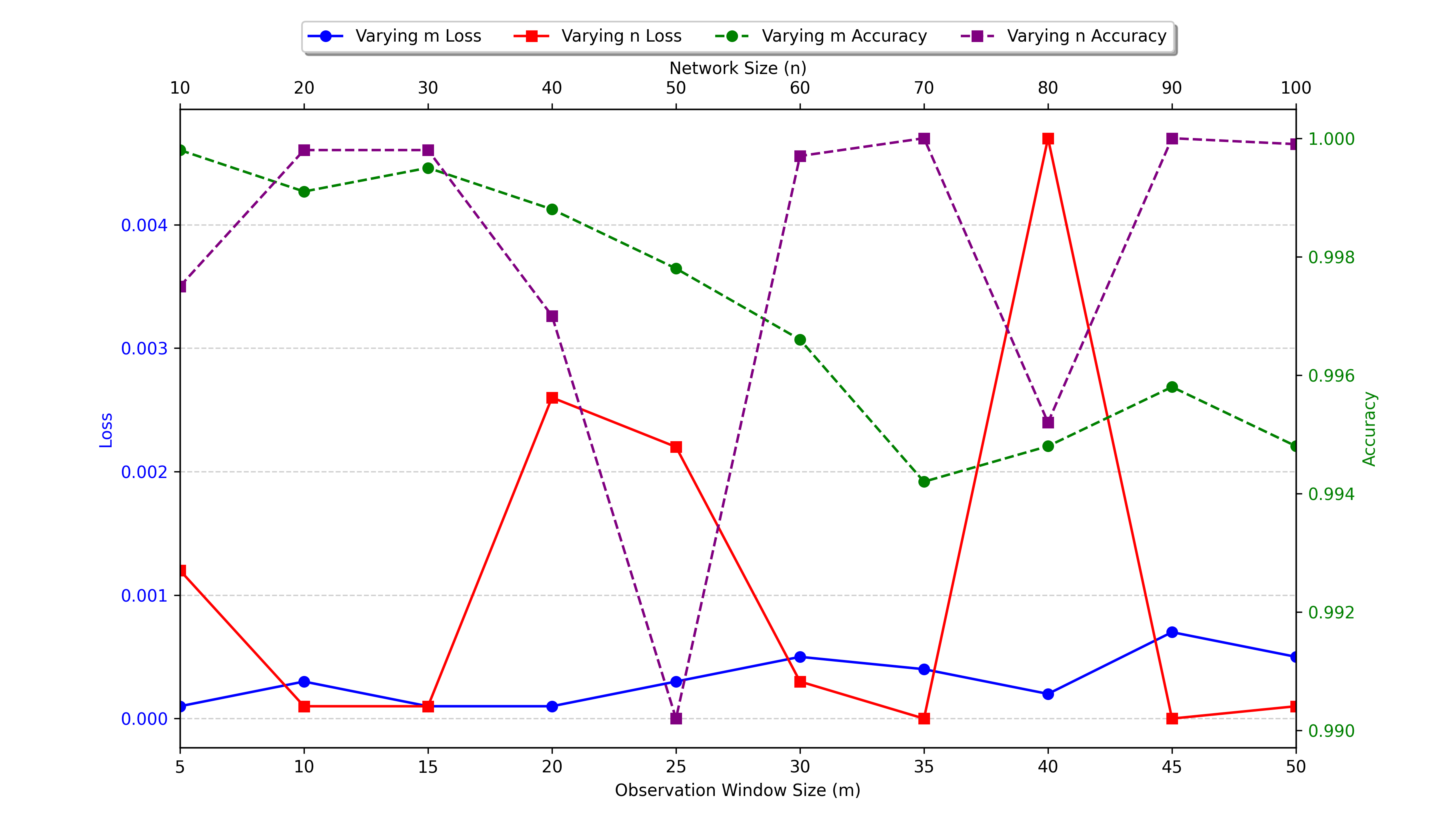}
    \caption{Model performance under varying $m$ and $n$.}
    \label{fig:performance_different_m_n}
\end{figure}

Figure~\ref{fig:performance_different_m_n} evaluates the impact of varying observation window size $m$ and network size $n$ using the global dataset. The network size was varied as $n \in \{10, 20, \dots, 100\}$, incremented by 10, while $m$ ranged from $\{5, 10, \dots, 50\}$ with a step size of 5. Across all configurations, the model maintained an accuracy of at least $0.990$ ($P_e = 0.01$), demonstrating its scalability and effectiveness for applications requiring larger observation windows or networks, such as intrusion detection in expansive communication networks, environmental monitoring in industrial sensor arrays, and others.

The behavior of Byzantine nodes as $m$ increases was examined to evaluate the duality reported in~\cite{MyInfoFusion2018}, where attackers alternate between $P_{mal}=0.5$ and $P_{mal}=1.0$ as the observation window grows. Larger windows enable malicious nodes to manipulate reports over time, with $P_{mal}=1.0$ maximizing errors through consistent flipping, while $P_{mal}=0.5$ introduces randomness that creates greater confusion and is harder for traditional fusion methods to counter. The maximum power strategy with $P_{mal}=1.0$ unintentionally reveals more information to the fusion center, making $P_{mal}=0.5$ a more challenging and attractive strategy for Byzantine nodes. Our deep learning model, trained on a comprehensive global adversarial dataset, effectively mitigates this duality by learning patterns across diverse attack scenarios. It adapts to both $P_{mal}=0.5$ and $P_{mal}=1.0$ strategies, achieving high accuracy and low error probabilities, thereby demonstrating robustness and strong generalization across varying adversarial conditions.

\subsubsection{Effect of Number of Samples Per Class}

\begin{figure}[!t]
    \centering
    \includegraphics[width=\columnwidth]{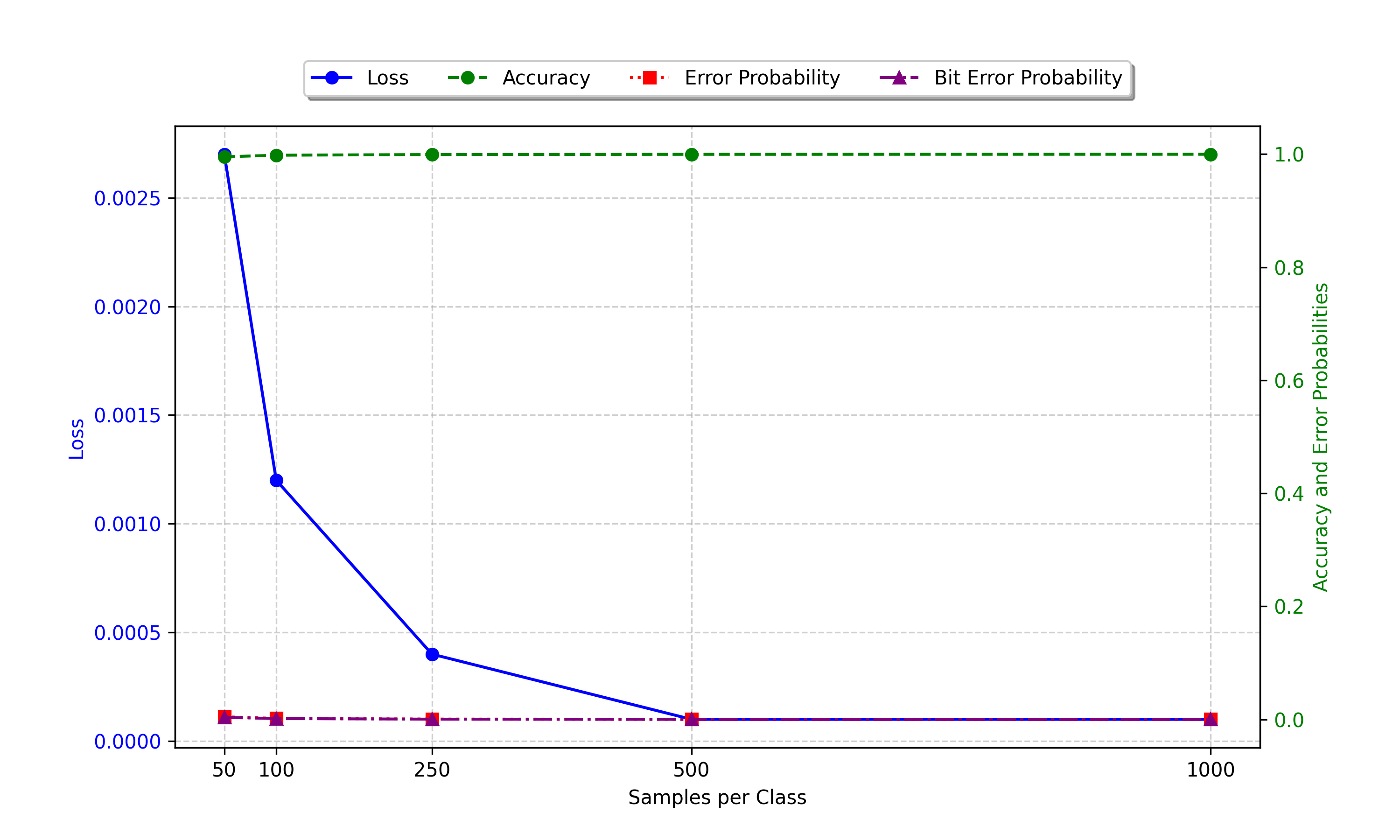}
    \caption{Model performance vs. number of samples per class.}
    \label{fig:acc_vs_nbsamples}
\end{figure}

Figure~\ref{fig:acc_vs_nbsamples} evaluates the effect of varying the number of samples per class on the model's performance using the global dataset. The results indicate a slight decrease in performance for $P_e$, accuracy, BER, and loss when the number of samples per class is below 100. Beyond this threshold, the performance stabilizes. We conclude that the model achieves high performance with $P_e$ close to or equal to 0.0 using as few as $200$ samples per class.

\subsubsection{Execution Time Analysis}
This experiment evaluates the computational efficiency of the proposed method by measuring training and inference times. The evaluation was conducted on the global dataset with $n=20$, $m=4$, and $200$ samples per class, resulting in a total dataset size of $15,200$ samples, divided into $80\%$ for training and $20\%$ for testing. The model was trained for $150$ epochs with a batch size of $512$. The total training time was $431.5482$ seconds, while the inference time for the entire test set was $0.1115$ seconds, corresponding to an average inference time of $0.00001048$ seconds per sample. Compared to state-of-the-art methods~\cite{MyTIFS, MyInfoFusion2018}, which require complex computations, this approach significantly reduces computational costs. Once trained, the method performs inference through a single forward propagation pass, making it highly suitable for real-time applications.

\section{Limitations and Discussion}
\label{sec.limitations}
While the proposed deep learning-based fusion approach demonstrates remarkable robustness and scalability across diverse adversarial scenarios, certain limitations should be acknowledged. First, the model's performance heavily relies on the quality and diversity of the training dataset. If the global adversarial dataset does not comprehensively cover the range of possible Byzantine behaviors or environmental conditions, the model may struggle to generalize effectively to unseen scenarios.

Second, although the model is trained offline on high-performance machines before being deployed to resource-constrained devices, the inference process on these devices can still impose computational overhead. For applications with extremely low-latency requirements or highly energy-efficient systems, optimizing the inference speed and memory usage of the model becomes critical to ensure seamless operation.

Finally, the interpretability of the proposed approach remains a challenge. While the model achieves high accuracy and robustness, understanding the specific decision-making processes within the neural network is difficult, which could possibly hinder trust and adoption in critical applications. Future work could explore methods to enhance interpretability and optimize the deployment process for constrained environments without compromising the model's robustness.

\section{Conclusions and Future Directions}
\label{sec.conc}
This paper introduced a deep learning-based approach for resilient decision fusion in adversarial multi-sensor networks, addressing the limitations of traditional methods that rely on specific assumptions and parameter tuning. We began by presenting a unified mathematical formulation to handle diverse adversarial scenarios, including varying proportions of Byzantine nodes, synchronized and unsynchronized attacks, unbalanced priors, adaptive strategies, and Markovian states. Subsequently, we proposed a deep neural network to replace the traditional fusion rules outlined in the unified framework. By training the neural network on a globally constructed dataset encompassing these scenarios, the method eliminates the need for scenario-specific models or prior knowledge of attack behaviors. Extensive experiments demonstrated the approach's robustness, achieving superior accuracy, minimal error probabilities, and adaptability to challenging conditions, such as Byzantine duality. The model's scalability, evidenced by its ability to effectively handle larger observation windows and network sizes, makes it suitable for applications such as intrusion detection, industrial sensor monitoring, and dynamic network security, all while maintaining computational efficiency for real-time use.

Future research can focus on extending the capabilities of the framework to address evolving and complex adversarial conditions. For instance, incorporating adversarial models that exploit access to observation vectors could provide a more comprehensive defense mechanism, as attackers with such access could selectively manipulate the most uncertain cases. Similarly, extending those adversarial models to scenarios where the system states follow a Markov sequence or are governed by other temporal dynamics could enhance its applicability in systems with memory-aware attacks. Exploring multi-class or multi-bit reporting schemes, as opposed to the binary decision fusion presented here, may also increase the robustness and versatility of the model in real-world applications.

Furthermore, optimizing the deployment of the model for edge devices and other resource-constrained environments could expand its practical applicability. Techniques such as model compression, quantization, or efficient inference strategies could address computational overhead and latency concerns without compromising accuracy. Finally, enhancing the interpretability of the neural network remains an important direction, as transparent decision-making processes could facilitate adoption in critical systems. This research highlights the transformative potential of deep learning in addressing the challenges of Byzantine-resilient decision fusion and sets the stage for further innovation in this domain.

\bibliographystyle{IEEEtran}
\bibliography{DLDecisionFusion}

\begin{thebibliography}{10}
\providecommand{\url}[1]{#1}
\csname url@samestyle\endcsname
\providecommand{\newblock}{\relax}
\providecommand{\bibinfo}[2]{#2}
\providecommand{\BIBentrySTDinterwordspacing}{\spaceskip=0pt\relax}
\providecommand{\BIBentryALTinterwordstretchfactor}{4}
\providecommand{\BIBentryALTinterwordspacing}{\spaceskip=\fontdimen2\font plus
\BIBentryALTinterwordstretchfactor\fontdimen3\font minus
  \fontdimen4\font\relax}
\providecommand{\BIBforeignlanguage}[2]{{%
\expandafter\ifx\csname l@#1\endcsname\relax
\typeout{** WARNING: IEEEtran.bst: No hyphenation pattern has been}%
\typeout{** loaded for the language `#1'. Using the pattern for}%
\typeout{** the default language instead.}%
\else
\language=\csname l@#1\endcsname
\fi
#2}}
\providecommand{\BIBdecl}{\relax}
\BIBdecl

\bibitem{WSNDDByz}
M.~Abdelhakim, L.~Lightfoot, J.~Ren, and T.~Li, ``Distributed detection in
  mobile access wireless sensor networks under byzantine attacks,'' \emph{IEEE
  Transactions on Parallel and Distributed Systems,}, vol.~25, no.~4, pp.
  950--959, April 2014.

\bibitem{WSNanomalyDet}
S.~Rajasegarar, C.~Leckie, M.~Palaniswami, and J.~C. Bezdek, ``Distributed
  anomaly detection in wireless sensor networks,'' in \emph{Proc. of ICCS 2006,
  10th IEEE Singapore International Conference on Communication systems}.\hskip
  1em plus 0.5em minus 0.4em\relax IEEE, 2006, pp. 1--5.

\bibitem{SS2010}
H.~Li and Z.~Han, ``Catching attacker(s); for collaborative spectrum sensing in
  cognitive radio systems: An abnormality detection approach,'' in \emph{IEEE
  Symposium on New Frontiers in Dynamic Spectrum, 2010}, April 2010, pp. 1--12.

\bibitem{Rawat11}
A.~S. Rawat, P.~Anand, C.~Hao, and P.~K. Varshney., ``Collaborative spectrum
  sensing in the presence of byzantine attacks in cognitive radio networks,''
  \emph{IEEE Transactions on Signal Processing}, vol.~59, no.~2, pp. 774--786,
  2011.

\bibitem{Zhang2013secure}
R.~Zhang, J.~Zhang, Y.~Zhang, and C.~Zhang, ``Secure crowdsourcing-based
  cooperative spectrum sensing,'' in \emph{Proc. of INFOCOM 2013, IEEE
  Conference on Computer Communications}.\hskip 1em plus 0.5em minus
  0.4em\relax IEEE, 2013, pp. 2526--2534.

\bibitem{Bar13}
M.~Barni and B.~Tondi, ``Multiple-observation hypothesis testing under
  adversarial conditions,'' in \emph{Proc. of WIFS'13, IEEE International
  Workshop on Information Forensics and Security}, Guangzhou, China, Nov 2013,
  pp. 91--96.

\bibitem{Lamp82}
L.~Lamport, R.~Shostak, and M.~Pease, ``The {B}yzantine generals problem,''
  \emph{ACM Trans. Program. Lang. Syst.}, vol.~4, no.~3, pp. 382--401, 1982.

\bibitem{Vemp13}
A.~Vempaty, T.~Lang, and P.~Varshney, ``Distributed inference with byzantine
  data: State-of-the-art review on data falsification attacks,'' \emph{IEEE
  Signal Processing Magazine}, vol.~30, no.~5, pp. 65--75, Sept 2013.

\bibitem{OptFusion}
Z.~Chair and P.~Varshney, ``Optimal data fusion in multiple sensor detection
  systems,'' \emph{IEEE Transactions on Aerospace and Electronic Systems}, vol.
  AES-22, no.~1, pp. 98--101, Jan 1986.

\bibitem{varshney2012distributed}
P.~K. Varshney, \emph{Distributed detection and data fusion}.\hskip 1em plus
  0.5em minus 0.4em\relax Springer Science \& Business Media, 2012.

\bibitem{MyBook}
A.~Abrardo, M.~Barni, K.~Kallas, B.~Tondi \emph{et~al.}, \emph{Information
  Fusion in Distributed Sensor Networks with Byzantines}.\hskip 1em plus 0.5em
  minus 0.4em\relax Springer, 2021.

\bibitem{Mar09}
S.~Marano, V.~Matta, and L.~Tong, ``Distributed detection in the presence of
  byzantine attacks,'' \emph{IEEE Transactions on Signal Processing,}, vol.~57,
  no.~1, pp. 16--29, 2009.

\bibitem{MyConsensus2016}
K.~Kallas, B.~Tondi, R.~Lazzeretti, and M.~Barni, ``Consensus algorithm with
  censored data for distributed detection with corrupted measurements: A
  game-theoretic approach,'' in \emph{Decision and Game Theory for Security:
  7th International Conference, GameSec 2016, New York, NY, USA, November 2-4,
  2016, Proceedings 7}.\hskip 1em plus 0.5em minus 0.4em\relax Springer, 2016,
  pp. 455--466.

\bibitem{tolerant_scheme}
R.~Chen, J.~M. Park, and K.~Bian, ``Robust distributed spectrum sensing in
  cognitive radio networks,'' in \emph{Proc. of INFOCOM 2008, 27th IEEE
  Conference on Computer Communications}, April 2008, pp.~--.

\bibitem{LearnByzantines}
A.~Vempaty, K.~Agrawal, P.~Varshney, and H.~Chen, ``Adaptive learning of
  byzantines' behavior in cooperative spectrum sensing,'' in \emph{Proc. of
  WCNC'11, IEEE Conf. on Wireless Communications and Networking}, March 2011,
  pp. 1310--1315.

\bibitem{ConditionalFrequencyCheck}
X.~He, H.~Dai, and P.~Ning, ``A byzantine attack defender: The conditional
  frequency check,'' in \emph{Information Theory Proceedings (ISIT), 2012 IEEE
  International Symposium on}, July 2012, pp. 975--979.

\bibitem{MyTIFS}
A.~Abrardo, M.~Barni, K.~Kallas, and B.~Tondi, ``A game-theoretic framework for
  optimum decision fusion in the presence of byzantines,'' \emph{IEEE
  Transactions on Information Forensics and Security}, vol.~11, no.~6, pp.
  1333--1345, 2016.

\bibitem{MyInfoFusion2018}
\BIBentryALTinterwordspacing
A.~Andrea, B.~Mauro, K.~Kassem, and T.~Benedetta, ``A message passing approach
  for decision fusion in adversarial multi-sensor networks,'' \emph{Information
  Fusion}, vol.~40, pp. 101--111, 2018. [Online]. Available:
  \url{https://www.sciencedirect.com/science/article/pii/S156625351630135X}
\BIBentrySTDinterwordspacing

\bibitem{MyMMEDIA2017}
\BIBentryALTinterwordspacing
A.~Abrardo, M.~Barni, K.~Kallas, and B.~Tondi, ``A message passing approach for
  decision fusion of hidden-markov observations in the presence of synchronized
  attacks,'' in \emph{International Conference on Advances in Multimedia},
  2017. [Online]. Available:
  \url{https://api.semanticscholar.org/CorpusID:196128961}
\BIBentrySTDinterwordspacing

\bibitem{MySynchronizedAttacksAPSIPA2018}
\BIBentryALTinterwordspacing
A.~Andrea, B.~Mauro, K.~Kassem, and T.~Benedetta, ``Decision fusion with
  unbalanced priors under synchronized byzantine attacks: a message-passing
  approach,'' \emph{2018 Asia-Pacific Signal and Information Processing
  Association Annual Summit and Conference (APSIPA ASC)}, pp. 1160--1167, 2018.
  [Online]. Available: \url{https://api.semanticscholar.org/CorpusID:71148973}
\BIBentrySTDinterwordspacing

\bibitem{ajay2022detection}
V.~Ajay and M.~Nesasudha, ``Detection of attackers in cognitive radio network
  using optimized neural networks.'' \emph{Intelligent Automation \& Soft
  Computing}, vol.~34, no.~1, 2022.

\bibitem{komar2016intelligent}
M.~Komar, A.~Sachenko, S.~Bezobrazov, and V.~Golovko, ``Intelligent cyber
  defense system using artificial neural network and immune system
  techniques,'' in \emph{International Conference on Information and
  Communication Technologies in Education, Research, and Industrial
  Applications}.\hskip 1em plus 0.5em minus 0.4em\relax Springer, 2016, pp.
  36--55.

\bibitem{behrens2023counter}
H.~W. Behrens, ``On counter-adversarial resilience in permeable networked
  systems,'' Ph.D. dissertation, Arizona State University, 2023.

\bibitem{MaxEntropyAttack}
Y.~Lin and H.~V. Zhao, ``Maximum entropy attack on decision fusion with herding
  behaviors,'' \emph{IEEE Signal Processing Letters}, vol.~31, pp. 2500--2504,
  2024.

\bibitem{MyCDC}
A.~Abrardo, M.~Barni, K.~Kallas, and B.~Tondi, ``Decision fusion with corrupted
  reports in multi-sensor networks: A game-theoretic approach,'' in \emph{53rd
  IEEE Conference on Decision and Control}.\hskip 1em plus 0.5em minus
  0.4em\relax IEEE, 2014, pp. 505--510.

\end{thebibliography}
\end{document}